\documentclass[final]{cvpr}

\usepackage{times}
\usepackage{epsfig}
\usepackage{graphicx}
\usepackage{amsmath}
\usepackage{amssymb}
\usepackage{lipsum,graphicx,subcaption}

\usepackage[pagebackref=true,breaklinks=true,colorlinks,bookmarks=false]{hyperref}

\graphicspath{{../}}

\def\cvprPaperID{27} 
\def\confYear{CVPR 2021}

\begin{document}

\title{Fine-Grained Visual Classification of Plant Species In The Wild: Object Detection as A Reinforced Means of Attention}

\author{
  Matthew~R.~Keaton
  \quad Ram~J.~Zaveri
  \quad Meghana~Kovur
  \quad Cole~Henderson
  \quad Donald~A.~Adjeroh \\
 Gianfranco~Doretto\thanks{This material is based upon work supported in part by the National Science Foundation under Grants No. OAC-1761792, IIS-1657179.}\\    
  West Virginia University, Morgantown, WV 26506\\
        \texttt{\small \{mrkeaton, rz0012, mk0174, cwh0015, daadjeroh, gidoretto\}@mix.wvu.edu } \\
      }
      

\maketitle

\begin{abstract}
Plant species identification in the wild is a difficult problem in part due to the high variability of the input data, but also because of complications induced by the long-tail effects of the datasets distribution. Inspired by the most recent fine-grained visual classification approaches which are based on attention to mitigate the effects of data variability, we explore the idea of using object detection as a form of attention. We introduce a bottom-up approach based on detecting plant organs and fusing the predictions of a variable number of organ-based species classifiers. We also curate a new dataset with a long-tail distribution for evaluating plant organ detection and organ-based species identification, which is publicly available\footnote{\url{https://github.com/wvuvl/DARMA}}.
\end{abstract}


\vspace{-4mm}
\section{Introduction}
\vspace{-1mm}

Automated plant image analysis in its many forms has become an increasingly relevant research topic, impacting several related areas of research and application~\cite{Soltis2020-bs}. Herbarium specimens have proven useful tools for phenological research~\cite{Willis2017-lu,Carranza-Rojas2017-xk,Das_Choudhury2019-jp}, while detection-based techniques are being deployed in precision agriculture~\cite{Mai2020-hl,Jiang2019-uy}. Furthermore, the collection of large crowdsourced datasets generated by ``citizen scientists''~\cite{Gura2013-fn} is expanding available research opportunities since image data is acquired ``in the wild,'' where factors including the quality, viewpoint, and illumination of images as well as the shape and scale of the plant subjects are fully unconstrained. While simultaneously providing larger pools of data and generating a more faithful representation of real-world scenarios, this type of data poses additional challenges in contrast with former applications where different degrees of control could be imposed on the way data is collected.


Settings in the wild increase the variability of the data, leading to a larger intra-class variance. The most recent approaches to fine-grained visual classification cope with that variability by leveraging an attention mechanism, whereby they focus on a subset of the available feature space with the hope of decreasing the sources of nuisance factors causing variation. Several successful approaches exist, including the use of specialized network layers~\cite{luo_cross-x_2019,ferrari_multi-attention_2018} and part-based attention~\cite{zheng_learning_2017,peng_object-part_2018, ferrari_hierarchical_2018, zheng_looking_2019}. On the other hand, when used in applications in the wild, where data often takes the form of a long-tailed distribution, these types of approaches tend to lose their effectiveness much more quickly as the number of samples per class decreases~\cite{Van_Horn2017-ul}.

The contribution of this work is twofold. First, we introduce a bottom-up approach to plant species identification in the wild that uses object detection as a means of attention to localize plant organs. This allows us to decrease the effects of data variability by basing identification on important regions of interest while ruling out unwanted background noise. At the same time, the supervised nature of the detection task (as opposed to unsupervised attention) allows for better mitigation of the long-tail effects still induced by settings in the wild.

The second contribution is the introduction of a long-tail dataset that allows for training a plant organ detector as well as plant organ-based species classifiers. To the best of our knowledge this is the first dataset of its kind, and it is publicly available. In the experiment section we evaluate the proposed approach on this new dataset.


\vspace{-2mm}
\section{Related work}
\vspace{-1mm}


{\bf Plant species identification.}
Classically, plant image analysis was constrained to species identification using leaves or flowers as image subjects, often consisting of a single leaf or cluster of leaves laid across a white background~\cite{wu_leaf_2007, novotny_leaf_2013, Soderkvist303038}. Other organs of interest have been analyzed, with best results stemming from the identification of flowers~\cite{arwatchananukul_new_2020, sfar_vantage_2013}. More recently, plant species identification has spread to more difficult and sizable datasets, increasing the number of available tasks in the field. Competitions, such as those hosted by ImageCLEF~\cite{Goeau2017-bo, Goeau2019-bq, Goeau2020-ex} and Google AI~\cite{tan_herbarium_2019, mwebaze_icassava_2019}, produce valuable datasets for various analysis tasks including species identification, expanding to broader open challenges such as fine-grained visual classification ``in the wild.'' Furthermore, crowdsourced datasets derived from images taken by ``citizen scientists,'' such as iNaturalist~\cite{van_horn_inaturalist_2018} and LeafSnap~\cite{leafsnap_eccv2012}, have gained popularity as their sheer size provides an advantage for deep neural networks.

{\bf Object detection.} Object detection has mostly had its use in plant analysis constrained to invasive species detection~\cite{goeau_plant_nodate} and phenotyping applications~\cite{buzzy_real-time_2020, Manacorda2018-wb}. This means related datasets either focus on detecting entire plants or individual organs, such as leaves, from a singular plant species. Recently, a few small-scale plant organ detection approaches and datasets using more than one species have been developed, each using at most a few hundred herbarium sheets with the intended use of phenological information gathering~\cite{Younis2020-vl,Ott2020-iu,Weaver2020-jy}. In the context of species identification, these approaches are unable to collect data on present biodiversity and the geographical distribution of different plant species, an advantage that crowdsourcing initiatives possess.



\vspace{-2mm}
\section{Proposed approach}
\vspace{-1mm}

\begin{figure}[t!]
\begin{center}
   \includegraphics[width=1.04\linewidth]{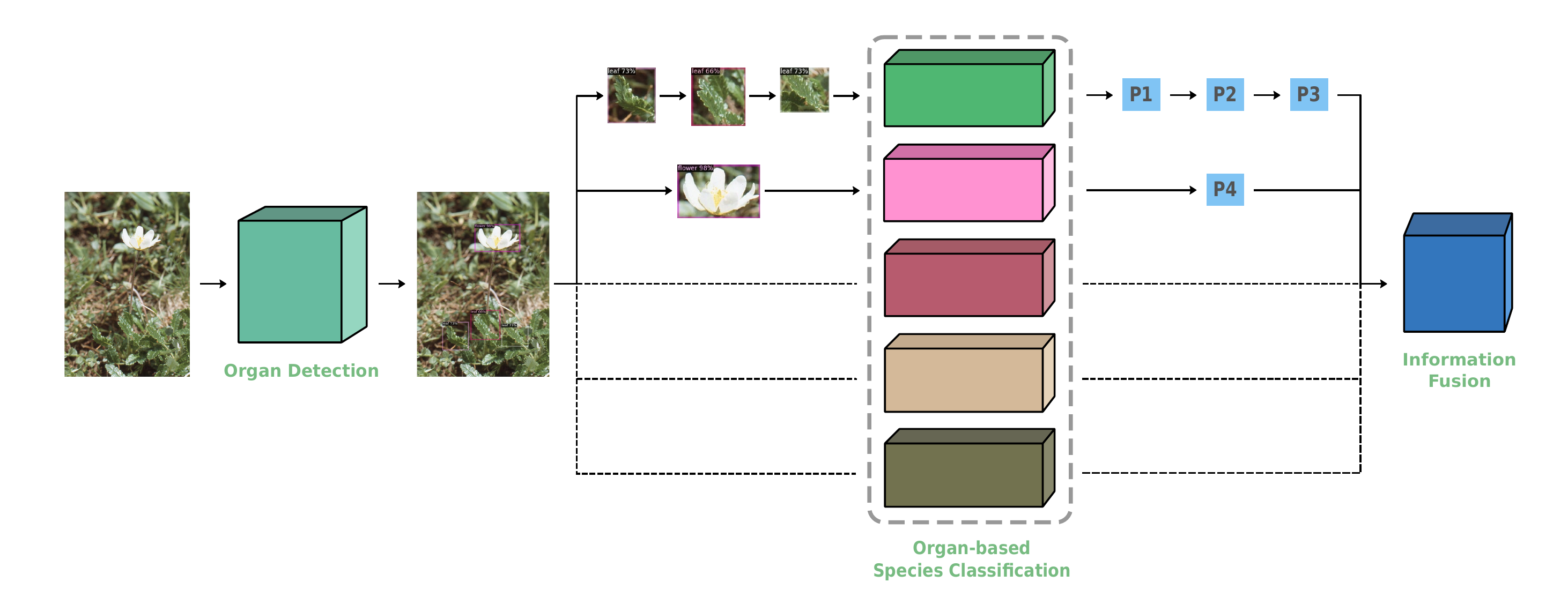}
 \end{center}
 \vspace{-6mm}
   \caption{\small{\textbf{Species identification in-the-wild overview.} The input image is processed by an organ detector producing the ROIs. Each ROI is fed into the corresponding} organ-based species classifier. Then, the per-organ species classification probabilities are aggregated by an information fusion step into the final species prediction.    }
\vspace{-5mm}
\label{fig:architecture}
\end{figure}

Our species identification approach contains three main components. First, an object detector identifies and localizes \emph{plant organs}, including leaves, flowers, fruit, stems, and regions with a high volume of leaves, termed ``high-density leaves'' (HDL). These regions of interest (ROIs) are then individually passed into an organ-based species classifier. 
Finally, the predictions for the given image are then aggregated by an information fusion step, which generates the final species prediction. See Figure~\ref{fig:architecture}.

Using an object detector to localize plant organs is advantageous for coping with the high variability of the data because it will select the information-rich regions to be individually classified while rejecting a large portion of the inherent background noise. Moreover, the downstream organ-based species classification and fusion steps will be able to handle a variable number of organs and to constructively aggregate predictions while being robust to false detections, as explained further below.

\textbf{Organ detection.}
Given an input image $I$, we obtain $n$ regions of interest for our downstream classifier by deploying an object detector, trained to localize and predict the classes of various plant organs. For the $i$-th ROI, the detector predicts the organ class $o_i$, where $o_i \in \{ \textsl{leaf, flower, fruit, stem, HDL} \}$. For this, we utilize the feature pyramid network-based Faster-RCNN object detector~\cite{Ren2017-le}, built on the ResNet-101 backbone, which is shown to attain strong baseline results even with a large variability of object sizes, as discussed in~\cite{Lin2017-ov}. Our model, pretrained on the COCO detection dataset, is built using the Detectron2 library~\cite{wu2019detectron2}. Using our dataset described in Section~\ref{sec:dataset}, we trained the model over $100$k iterations using an SGD optimizer with a base learning rate of $5\times 10^{-5}$ and momentum of 0.9. Additionally, a non-maximum suppression threshold of $0.1$ is used during both training and testing. Due to memory constraints, a ``mini-batch'' of $1$ is used; training takes approximately 10 hours when run on a single NVIDIA GTX 1660 GPU.






\textbf{Organ based species classification.}
Given the $i$-th ROI labeled by the organ detector as $o_i$, we compute the probability of the ROI to depict the species $s$, $p(s|o_i)$. We do so with an organ-based species classifier implemented with a convolutional neural network ending with a softmax layer. We use a ResNet-18~\cite{he2016deep} as the backbone network, which we train with a cross-entropy loss. Each ROI was resized to $224 \times 224$, which is the average ROI size of the dataset. Additionally, each classifier was fine-tuned from a model pre-trained on ImageNet, for just enough iterations to observe the validation loss reaching the plateau. We used the default Adam optimizer, and learning rate of $1\times 10^{-4}$, with a minibatch of size 32.

\begin{figure*}[th]
\begin{center}
     \includegraphics[width=0.75\linewidth]{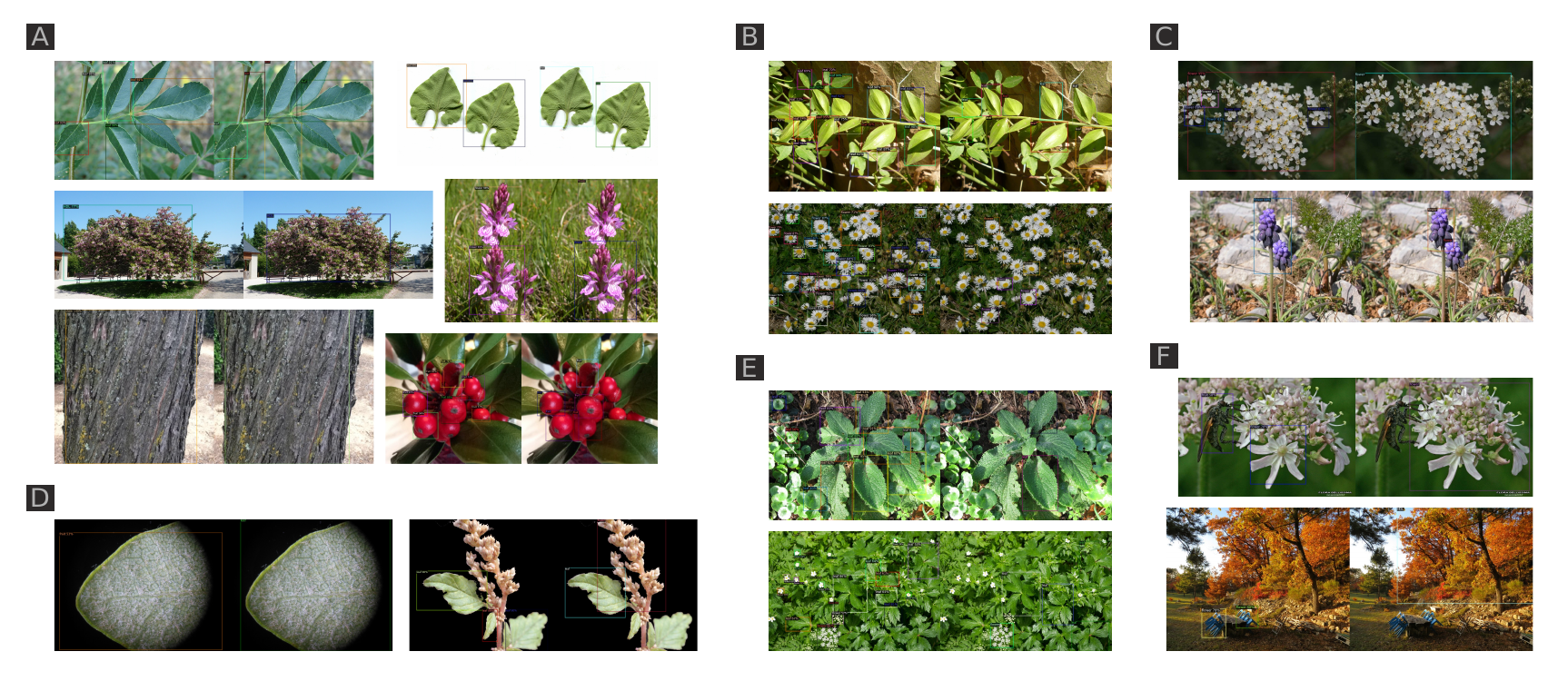}
   \end{center}
   \vspace{-8mm}
   \caption{\small{\textbf{Sample annotations from the DARMA dataset.} In (a), clockwise from the top-left, are samples of leaves, ``scan-like'' leaves that make up approximately an eighth of the dataset, flowers, fruit, stems and HDL. Highlighted in (b) through (f) are various challenges that potentially impact organ detection or classification performance. (b) and (c) depict artificial losses in average precision due to only a subset of leaves being annotated in (b) and random grouping/non-grouping of flowers due to annotator preferences in (c). (d)-(f) show various difficulties from the dataset and approach that affect classification accuracy. (d) shows random image effects added to photographs by submitters of the images, (e) provides examples where non-target species are included in an image and then extracted by the organ detector, and (f) shows that vibrant or large non-plant items found in images can confuse the organ detector.}}
\label{fig:annotation_figure}
   \vspace{-4mm}
\end{figure*}

\textbf{Information fusion.}
The species prediction entails finding the species $s$ that maximizes the probability $p(s)$. We can express $p(s)$ as $p(s)=\sum_i p(s|o_i)p(o_i)$, where $p(o_i)$ is the prior on the organ $o_i$, and each of the organs in the input image are assumed to be independent. In lack of prior knowledge, the natural choice is to assume a uniform prior, which means that $p(s)$ becomes the average of the probabilities $p(s|o_i)$. This is equivalent to the so-called \emph{sum rule} in information fusion~\cite{fusion_sum_vote}.

Another way to pose the species prediction problem is to find the species $s$ that maximizes the posterior $p(s|o_1, \cdots, o_n)$. Under the assumption that every species is equally likely, and that the organs are independent conditioned on the species, this is equivalent to finding $s$ that maximizes $\prod_i p(o_i|s)$, which is the naive Bayesian fusion. If we further assume a uniform prior on the organs (like above), this is equivalent to finding $s$ that maximizes $\prod_i p(s|o_i)$. This is known as the \emph{product rule} in information fusion~\cite{Dubois2016-ur}.

Finally, when $p(s|o_i)$ is approximated by a one-hot vector with the 1 corresponding to the species that maximizes $p(s|o_i)$ the sum rule becomes the \emph{voting rule}~\cite{fusion_sum_vote}. In the experiments we compared both the sum, the product, and the voting rule.

\vspace{-2mm}
\section{DARMA dataset} \label{sec:dataset}
\vspace{-1mm}

In order to demonstrate the effectiveness of our approach, we first cultivated a benchmark dataset due to the lack of one for organ detection in the wild, which we named DARMA (i.e., short for Detection as A Reinforced Means of Attention). We derive our images from the Pl@ntView dataset used in the PlantCLEF 2015 challenge~\cite{Joly2015-iw}. The images were gathered as part of a citizen scientist initiative, meaning that a large portion of them were taken by amateur photographers (see Figure~\ref{fig:annotation_figure}).

Our dataset separates itself from the PlantCLEF 2015 dataset with the addition of bounding box annotations for four different plant organs - leaves, flowers, fruit, and stems - and regions we call high-density leaves (HDL). Additionally, the PlantCLEF 2015 challenge was centered around multi-observation queries, where one to five images are provided as part of a singular instance for prediction. Instead, we split these queries and always provide only one image for each prediction.
\begin{figure}[t]
\begin{center}
   \includegraphics[width=0.48\linewidth]{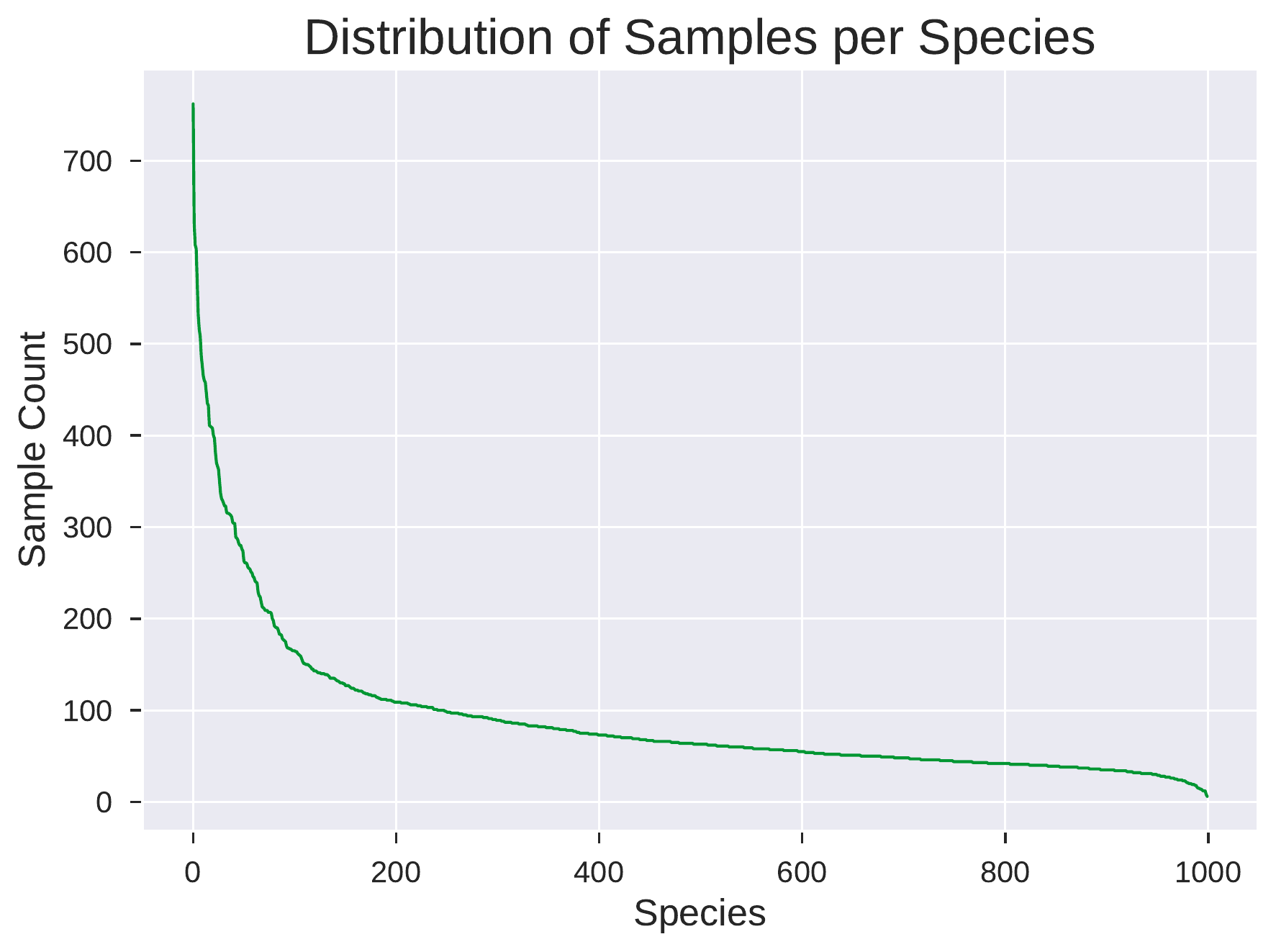}
   \includegraphics[width=0.48\linewidth]{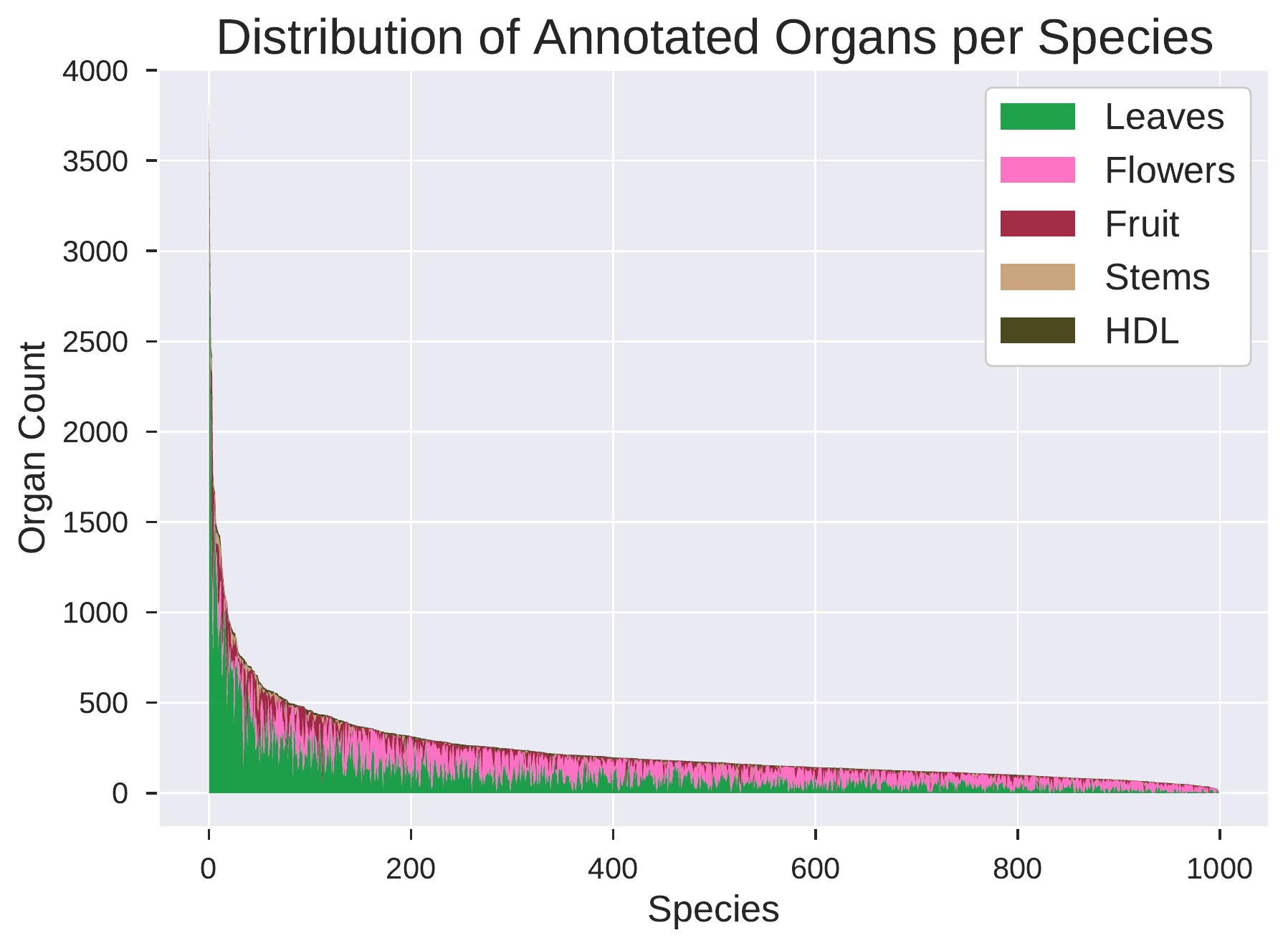}
 \end{center}
 \vspace{-5mm}
   \caption{\label{fig:distribution}
\small{\textbf{Distribution of samples (left) and annotated organs (right) across each species.} Species are sorted in descending order of sample and total annotated organ count; both the sample number and organ count follow a long-tailed distribution.}}
 \vspace{-3mm}
\end{figure}

\begin{table}
\begin{center}\footnotesize
\begin{tabular}{|c|c|c|c|}
\hline
Train & Test & Validation \\
\hline\hline
62959 & 17995 & 9016 \\
\hline
\end{tabular}
\end{center}
\vspace{-5mm}
\caption{\label{tab:samples} \small{\textbf{Statistics on the number of samples per split.}}}
\vspace{-3mm}
\end{table}

\begin{table}
\begin{center}\footnotesize
\begin{tabular}{|c|c|c|c|}
\hline
Organs & Average & Standard Deviation \\
\hline\hline
Leaf & 184$\times$199 & 145$\times$172 \\
Flower & 210$\times$220 & 188$\times$190 \\
Fruit & 141$\times$151 & 142$\times$146 \\
Stem & 516$\times$610 & 239$\times$220 \\
HDL & 634$\times$565 & 167$\times$155 \\
\hline
\end{tabular}
\end{center}
\vspace{-5mm}
\caption{\label{tab:bounding-boxes}\small{\textbf{Statistics on the scale of bounding boxes per organ.}}}
\vspace{-3mm}
\end{table}

For each species, the images were split into 70\% for training, 10\% for validation, and 20\% for testing. When fewer than 10 images were included for a species, at least one sample was placed into the validation and test set. Annotations were generated following a specific protocol. For leaves, only those that were normal to the image plane with less than 25\% of their surface obstructed were to be annotated. Flowers following the same guideline for obstruction were annotated as well, including flower buds. The fruit category included fruits as well as pine cones and seeds. Stems included both upright (mostly vertical) plant stems and tree bark. Finally, we introduce the HDL category. These are regions where leaves are difficult to differentiate due to the distance the photo was taken from, and they include information in the form of a texture rather than a shape. Tables~\ref{tab:samples},~\ref{tab:bounding-boxes},~\ref{tab:samples-species}, and~\ref{tab:organs} illustrate the basic statistics of the dataset.

\begin{figure*}[th]
\begin{center}
 \includegraphics[width=0.19\linewidth]{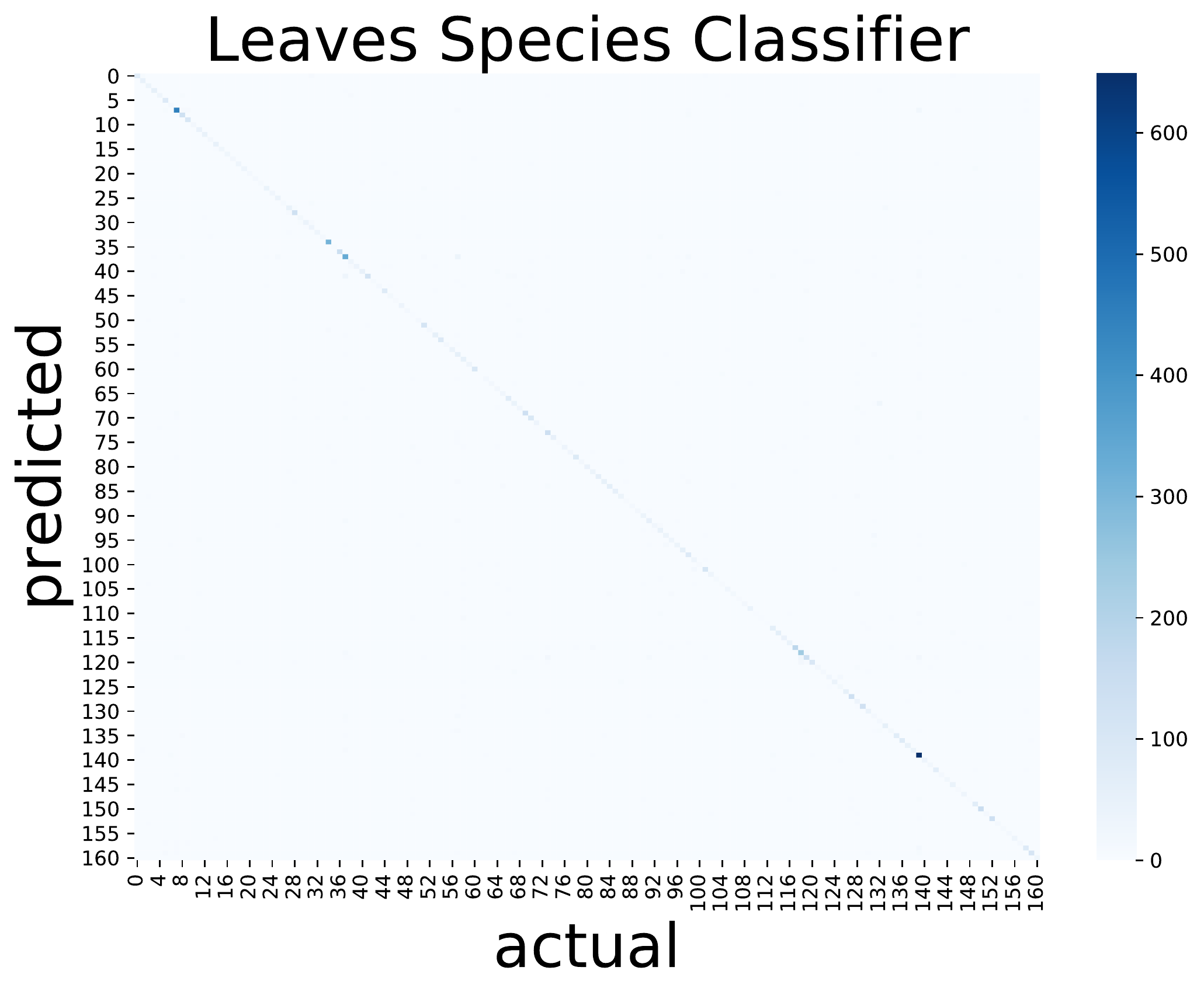}
 \includegraphics[width=0.19\linewidth]{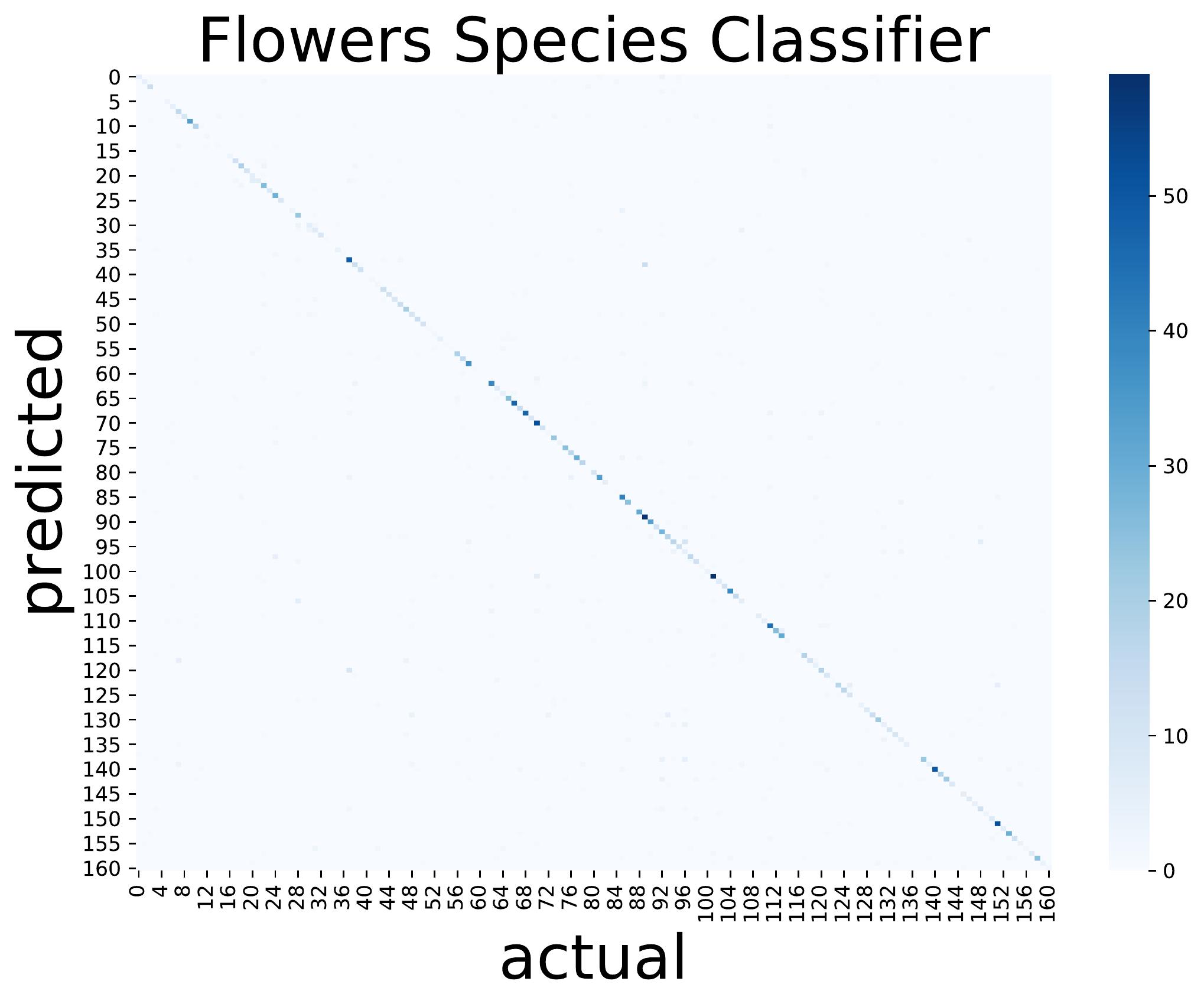}
 \includegraphics[width=0.19\linewidth]{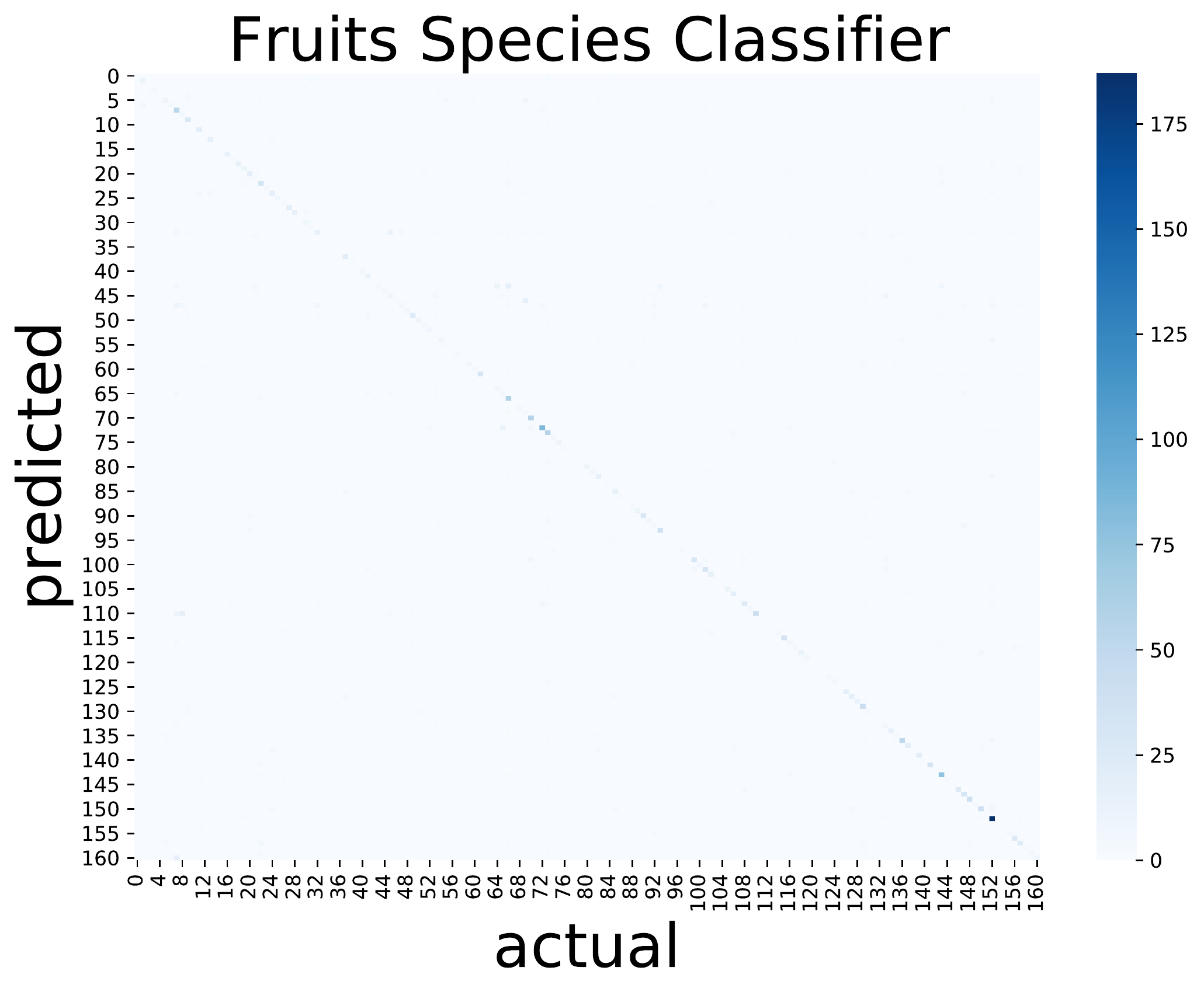}
 \includegraphics[width=0.19\linewidth]{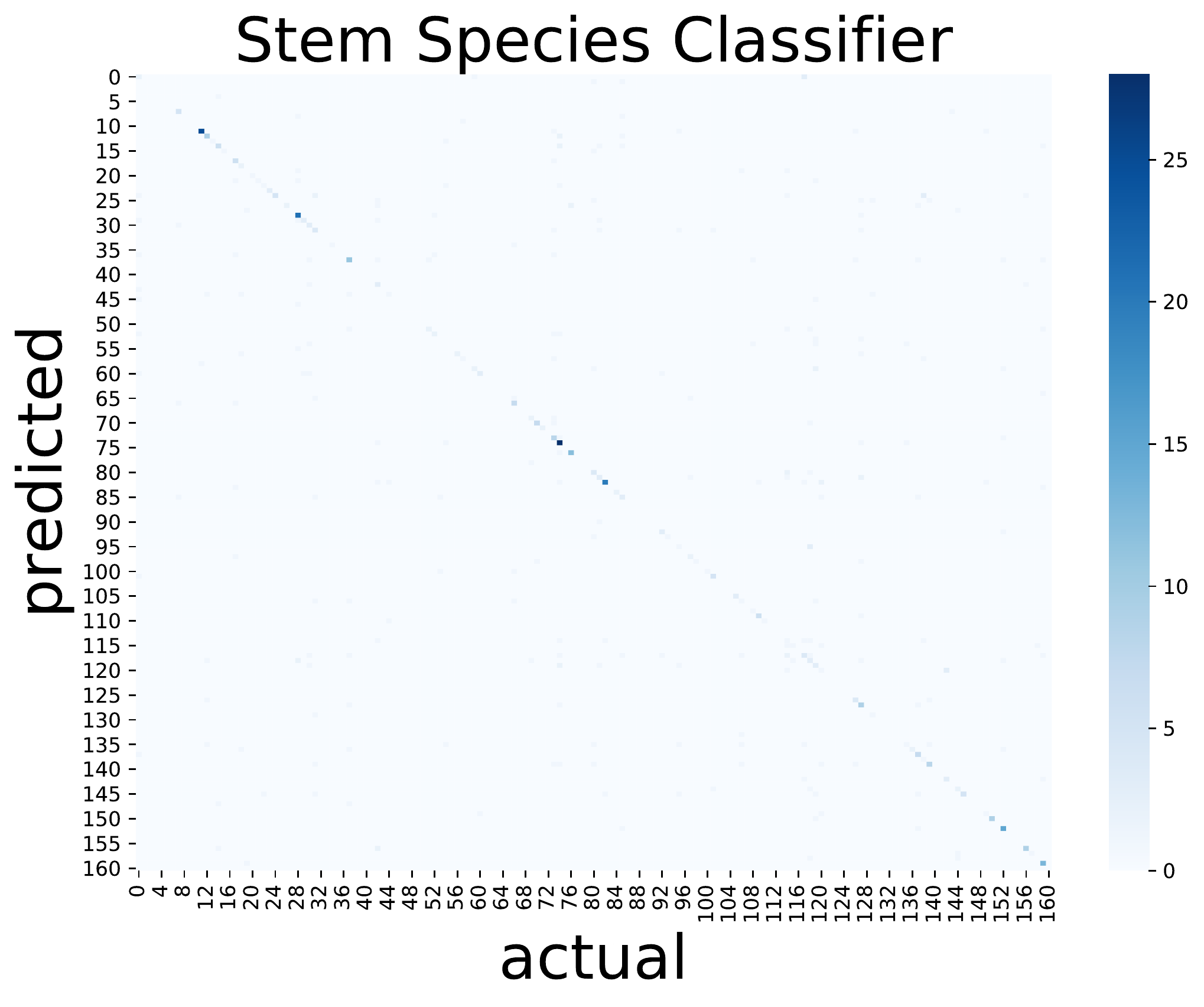}
 \includegraphics[width=0.19\linewidth]{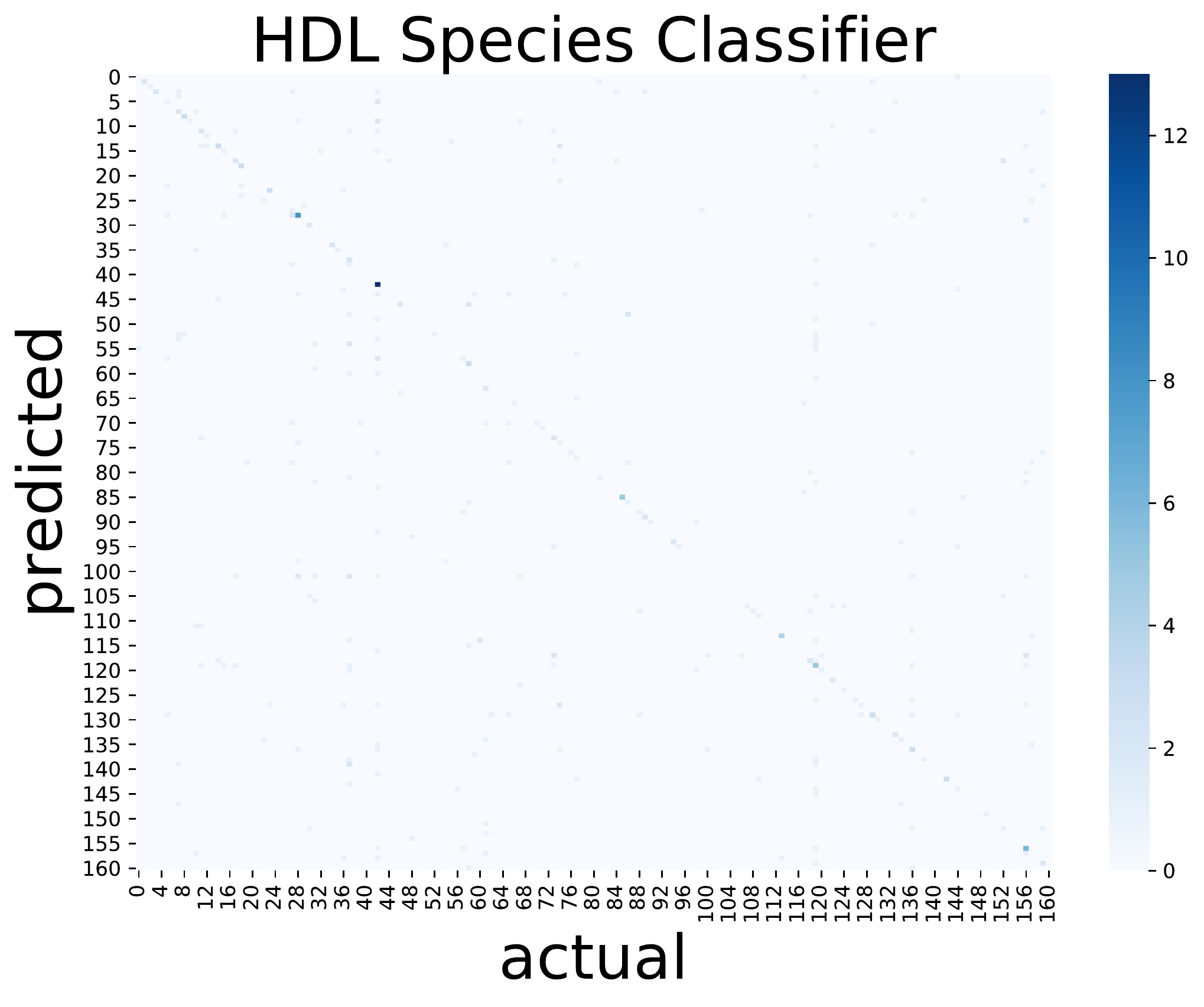}
\end{center}
\vspace{-5mm}
\caption{\small{\textbf{Organ-based species classification.} Confusion matrices for each of the organ-based classifiers.}}
\label{fig:confusion-matrices}
\vspace{-3mm}
\end{figure*}

Besides the images being taken in the wild, the dataset presents additional challenges. Figure~\ref{fig:distribution} shows that the distribution of both images and organ annotations across species exhibit long tails where a large proportion of samples and organs is possessed by a few classes. In terms of each image, there are often other plants that appear within the frame, making proper classification more difficult. In addition, not all organs are annotated within each image - annotators were instructed to include as many as three leaves when at least that many were present. Although this should have a small impact on the efficacy of the object detector, AP results may be artificially lowered because of this, and efforts to improve annotations will be made in the future. Figure~\ref{fig:annotation_figure} portrays several samples with annotations and associated challenges.

\vspace{-2mm}
\section{Experiments}
\vspace{-1mm}

\textbf{Organ detection.}
Our initial results are outlined in Tables~\ref{tab:detection-ap} and~\ref{tab:organ-ap}. The organ detector tends to perform lower on our dataset than on other object detection datasets; this is due to a number of reasons, as discussed in Section~\ref{sec:dataset}. However, it is important to note that the overall approach is somewhat robust with respect to the calculated performance of the organ detector. This is because a few false positives are likely filtered out by the fusion stage of the final prediction. Moreover, several unannotated organs are also picked up by the detector, reinforcing a correct final species prediction while decreasing the average precision of the detector.


\begin{table}
\begin{center}\footnotesize
\begin{tabular}{|c|c|c|c|}
\hline
Mean & Standard Deviation & Minimum & Maximum \\
\hline\hline
90.0 & 85.6 & 6 & 762 \\
\hline
\end{tabular}
\end{center}
\vspace{-5mm}
\caption{\label{tab:samples-species} \small{\textbf{Statistics on the number of samples per species.}}}
\vspace{-3mm}
\end{table}

\begin{table}
\begin{center}\footnotesize
\begin{tabular}{|l|c|c|c|}
\hline
Organ & Mean & Standard Deviation & Maximum \\
\hline\hline
Leaf & 119.0 & 219.7 & 3568 \\
Flower & 82.7 & 67.0 & 606 \\
Fruit & 30.7 & 66.1 & 1122 \\
Stem & 4.4 & 14.6 & 153 \\
HDL & 5.2 & 7.8 & 94 \\
\hline
\end{tabular}
\end{center}
\vspace{-5mm}
\caption{\label{tab:organs} \small{\textbf{Statistics on the number of organs per species.} Minimum values are 0 for all organs.}}
\vspace{-3mm}
\end{table}

\begin{table}
\begin{center} \footnotesize
\begin{tabular}{|c|c|c|}
\hline
$AP$ & $AP_{50}$ & $AP_{75}$ \\
\hline\hline
42.9 & 66.6 & 46.2 \\
\hline
\end{tabular}
\end{center}
\vspace{-5mm}
\caption{\small{\textbf{Organ detection average precision.} We use the COCO AP evaluation metrics. The first AP is averaged across threshold values ranging from 0.50 to 0.95 with increments of 0.05.}}
\label{tab:detection-ap}
\vspace{-3mm}
\end{table}

\begin{table}
\begin{center} \footnotesize
\begin{tabular}{|c|c|c|c|c|}
\hline
Leaf & Flower & Fruit & Stem & HDL \\
\hline\hline
40.9 & 36.0 & 25.7 & 74.0 & 37.6 \\
\hline
\end{tabular}
   \vspace{-5mm}
\end{center}
\caption{\small{\textbf{Average precision for each organ.} AP is calculated in the same manner as in Table~\ref{tab:detection-ap}.}}
\label{tab:organ-ap}
   \vspace{-3mm}
\end{table}

\begin{table}
  \begin{center} \footnotesize
\begin{tabular}{|c|c|c|c|c|}
\hline
Leaf & Flower & Fruit & Stem & HDL \\
\hline\hline
68.24 & 75.24 & 63.39 & 58.24 & 34.21 \\
\hline
\end{tabular}
\end{center}
\vspace{-5mm}
\caption{\small{\textbf{Organ-based classification accuracy.}}}
\label{tab:organ-accuracy}
\vspace{-3mm}
\end{table}

\textbf{Organ-based species classification.} We down-selected 161 species out of the original 1000 to retain those that had ROI data for every organ, and that had at least 130 leaf samples. Table~\ref{tab:organ-accuracy} reports the accuracy of each of the organ-based classifiers. There is a significant accuracy drop for stem and HDL, which was to be expected given the much lower availability of data for these organs. In addition, Figure~\ref{fig:confusion-matrices} shows the confusion matrices.

\begin{table}
\begin{center} \footnotesize
\begin{tabular}{|c|c|c|c||c|} 
\hline
Rule & Sum & Product & Voting & ResNet-18 \\
\hline\hline
Accuracy & 79.36 & 78.16 & 77.65 & 69.65 \\
\hline
\end{tabular}
\end{center}
\vspace{-5mm}
\caption{\small{\textbf{Species identification accuracy.} Comparison among three fusion techniques: sum rule, product rule, and voting rule.}}
\vspace{-3mm}
\label{tab:fusion}
\end{table}

\textbf{Fusion-based species identification.}
Table~\ref{tab:fusion} shows the species identification accuracy for the three fusion approaches, where the predictions were made on a per-image basis. The sum rule appears to outperform the others. In particular, we observe that the fusion accuracy outperforms the organ-based accuracies, proving the efficacy of combining multiple observations of discriminatory features from a single image. Finally, the fusion accuracy outperforms also the baseline of 69.65 obtained by training ResNet-18 on the whole input image for 30 epochs. 




\vspace{-2mm}
\section{Conclusions}
\vspace{-1mm}

We introduced an approach for plant species identification in the wild and a new dataset for evaluating organ detection and organ-based species identification. We have shown that the dataset is long-tail distributed. Based on our initial evaluation, the approach exhibits robustness against false detections by fusing multiple species predictions, achieving accuracy values comparable with top-performing entries from related challenges in the wild.

{\small
\bibliographystyle{ieee_fullname}
\bibliography{../references}
}

\end{document}